\documentclass[10pt,twocolumn,letterpaper]{article}

\usepackage[pagenumbers]{wacv} % To force page numbers, e.g. for an arXiv version

\usepackage[accsupp]{axessibility}  % Improves PDF readability for those with disabilities.
\usepackage{graphicx}
\usepackage{amsmath}
\usepackage{amssymb}
\usepackage{booktabs}
\usepackage{multirow}
\usepackage[table]{xcolor}
\usepackage[pagebackref,breaklinks,colorlinks]{hyperref}

\usepackage[capitalize]{cleveref}
\crefname{section}{Sec.}{Secs.}
\Crefname{section}{Section}{Sections}
\Crefname{table}{Table}{Tables}
\crefname{table}{Tab.}{Tabs.}

\def\wacvPaperID{637} % *** Enter the WACV Paper ID here
\def\confName{WACV}
\def\confYear{2024}

\begin{document}

\title{StreamMapNet: Streaming Mapping Network for Vectorized Online HD Map Construction}

% \author{Tianyuan Yuan\\
% Tsinghua University\\
% {\tt\small yuanty22@mails.tsinghua.edu.cn}
% \and
% Yicheng Liu\\
% Tsinghua University\\
% \and
% Yicheng Liu\\
% Tsinghua University\\
% \and
% Yicheng Liu\\
% Tsinghua University\\
% \and
% Yicheng Liu\\
% Tsinghua University\\

\author{Tianyuan Yuan$\phantom{}^{1}$\hspace{15pt}
Yicheng Liu$\phantom{}^{1}$\hspace{15pt}
Yue Wang$\phantom{}^{2}$\hspace{15pt}
Yilun Wang$\phantom{}^{1}$\hspace{15pt}
Hang Zhao$\phantom{}^{1}$\thanks{Corresponding at: \texttt{hangzhao@mail.tsinghua.edu.cn}} \vspace{8pt}\\
$\phantom{}^1$Tsinghua University\hspace{5pt}
$\phantom{}^2$University of Southern California\hspace{5pt}
}
\maketitle

%%%%%%%%% ABSTRACT
\begin{abstract}
High-Definition (HD) maps are essential for the safety of autonomous driving systems. While existing techniques employ camera images and onboard sensors to generate vectorized high-precision maps, they are constrained by their reliance on single-frame input. This approach limits their stability and performance in complex scenarios such as occlusions, largely due to the absence of temporal information. Moreover, their performance diminishes when applied to broader perception ranges.
In this paper, we present StreamMapNet, a novel online mapping pipeline adept at long-sequence temporal modeling of videos. StreamMapNet employs multi-point attention and temporal information which empowers the construction of large-range local HD maps with high stability and further addresses the limitations of existing methods.
Furthermore, we critically examine widely used online HD Map construction benchmark and datasets, Argoverse2 and nuScenes, revealing significant bias in the existing evaluation protocols. We propose to resplit the benchmarks according to geographical spans, promoting fair and precise evaluations.
Experimental results validate that StreamMapNet significantly outperforms existing methods across all settings while maintaining an online inference speed of $14.2$ FPS. 
% Our code will be made publicly available.
Our code is available at \url{https://github.com/yuantianyuan01/StreamMapNet}.
\end{abstract}

%%%%%%%%% BODY TEXT
\section{Introduction}
\label{sec:intro}
High-Definition (HD) maps, designed specifically for autonomous driving, are highly accurate maps that provide detailed and vectorized representations of map elements such as pedestrian crossings, lane dividers, and road boundaries. These maps are essential for self-driving vehicles as they contain rich semantic information about roads, enabling effective navigation.
Traditionally, HD maps were constructed offline using SLAM-based methods (LOAM~\cite{loamzhang2014}, LeGO-LOAM~\cite{legoloam2018}), resulting in complex pipelines and high maintenance costs. However, these methods face scalability issues due to their heavy reliance on human labor for map annotation and updates. In recent years, deep-learning-based methods have emerged as a promising alternative, allowing for the online construction of vectorized HD maps around the ego-vehicle using onboard sensors. These online approaches offer cost savings by eliminating the need for mapping fleets and reducing human labor while maintaining the ability to adapt to new environments or potential map changes.

Early approaches to semantic map learning treated the task as a segmentation problem in Bird's-Eye-View (BEV) space (HDMapNet~\cite{li2021hdmapnet}, Lift-Splat-Shoot~\cite{philion2020lift}, Roddick and Cipolla~\cite{roddick2020predicting}). However, these methods generated rasterized maps that lacked the notion of instances, rendering them unsuitable for downstream tasks such as motion forecasting or motion planning (VectorNet~\cite{vectornet}, LaneGCN~\cite{liang2020learning}) which require vectorized maps. More recently, approaches like VectorMapNet~\cite{liu2023vectormapnet} and MapTR~\cite{MapTR} achieved promising results in constructing end-to-end vectorized local HD maps using transformer~\cite{vaswani2017attention} decoders inspired by DETR~\cite{carion2020endtoend}. Nevertheless, these methods face two main challenges: \textit{(1) Small perception range:} These methods are limited in constructing HD maps with a relatively small perception range of $60\times 30\,m$, which is impractical for autonomous driving scenarios. When attempting to extend the perception range to a larger scale, such as $100\times 50\,m$, their performance significantly deteriorates. \textit{(2) Not leveraging temporal information:} These approaches only leverage single-frame inputs and fail to exploit temporal information. As a result, these methods are prone to errors caused by challenging environmental conditions such as occlusions, large crossroads, and extreme camera exposures, which are quite frequent in autonomous driving scenarios. Additionally, temporal inconsistency between maps of different timestamps is extremely challenging for the motion planning module as it creates a constantly changing world for autonomous driving system. To address these issues, we present StreamMapNet, an end-to-end online pipeline that utilizes camera videos with a temporal fusion strategy to construct temporal-consistent high-quality vectorized maps covering a wide range.

We frame the map construction task as a detection problem. Our model adopts an encoder-decoder architecture: a general BEV encoder that aggregates features from multiple-view images, and a DETR-like decoder for decoding map element instances.
Specifically, we assign one object query to associate with each map element. 
Unlike typical object detection scenarios where objects exhibit local characteristics, map elements often possess irregular and elongated shapes, necessitating long-range attention modeling that conventional deformable attention~\cite{zhu2020deformable} fails to capture. To enable a wide perception range, we introduce a novel approach called ``Multi-Point Attention" that effectively captures longer attention ranges while maintaining computational efficiency.

\begin{figure}[t]
  \centering
   \includegraphics[width=1.0\linewidth]{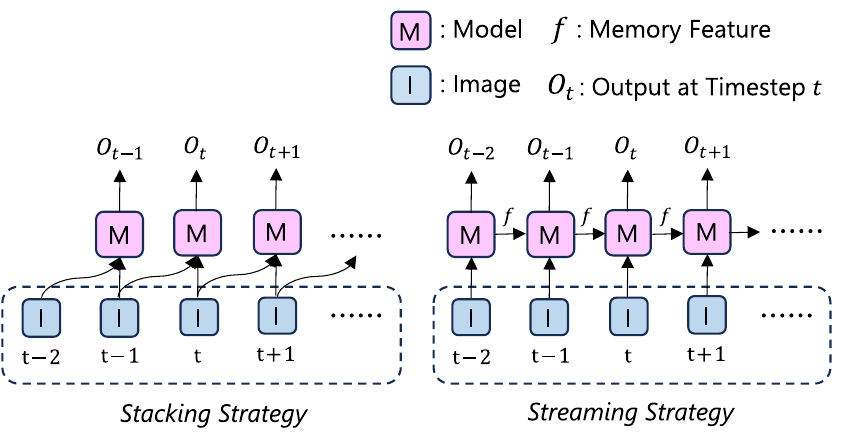}
   \caption{Comparison between \textit{stacking} strategy and \textit{streaming} strategy. \textit{streaming} strategy encodes all historical information into the memory feature to save cost and build long-term association. }
   \label{fig:streaming}
\end{figure}

We adopt a \textit{streaming} strategy for our temporal fusion approach. Instead of \textit{stacking} multiple frames together, our method process each frame individually while propagating a hidden state across time to preserve temporal information, as demonstrated in Figure \ref{fig:streaming}. This \textit{streaming} strategy offers two key advantages over the \textit{stacking} strategy: (1) it facilitates longer temporal associations as the propagated hidden states encode all historical information, and (2) it minimizes memory and latency costs compared to the \textit{stacking} strategy, which consumes memory and computational resources linearly with the number of stacked frames. Recent works in 3D object detection share the same spirit (VideoBEV~\cite{han2023exploring}, StreamPETR~\cite{wang2023exploring}, Sparse4D v2~\cite{lin2023sparse4d}).
In our framework, the propagated hidden states encompass BEV features and refined object queries. We design a dedicated temporal fusion module for each of these components.

% Finally, we critically examine the current evaluation setting utilized in the nuScenes dataset, which has been employed by recent online vectorized map construction tasks such as VectorMapNet, MapTR, and BeMapNet. We assert that there are significant fairness issues within the evaluation setting arising from an inappropriate training-validation split. Specifically, we discover that over 84\% of the locations in the validation split are also present in the training split, leading to overfitting problems. A similar concern has been identified in the Argoverse2 dataset. In order to establish a more equitable benchmark for this task, we introduce new training-validation splits for both datasets, ensuring there is no overlap between the splits. By doing so, we aim to provide a fairer basis for future research in this field. For fair comparison, we conduct comprehensive experiments on both the \textit{old} and \textit{new} splits for both datasets. The quantitative results demonstrate that our method outperforms all existing state-of-the-art approaches across \textbf{all} experimental settings.

Lastly, we critically examine the current evaluation setup in the nuScenes dataset~\cite{caesar2020nuscenes}, a common benchmark for recent online vectorized map construction methods such as VectorMapNet~\cite{liu2023vectormapnet}, MapTR~\cite{MapTR}, and BeMapNet~\cite{qiao2023endtoend}. Our investigation reveals substantial fairness issues in this setup due to a problematic training-validation split. Specifically, we find that more than $84\%$ of validation locations are also present in the training split, which could lead to overfitting. We identified a similar issue with the Argoverse2 dataset~\cite{Argoverse2}.
In response, we propose using new, non-overlapping training-validation splits for both datasets, aiming to build a fairer benchmark for future research in this area. To ensure an equitable comparison, we perform extensive experiments on both the original and new splits for each dataset. The resulting quantitative results confirm the superior performance of our method across all experimental settings, surpassing all existing state-of-the-art approaches.

To summarize, our contributions are as follows:
\begin{itemize}
    \item We introduce a novel approach called "Multi-Point Attention" to extend the perception range of local vectorized HD maps to $100\times 50$ meters, demonstrating improved practicality without experiencing a significant performance drop.
    \item We design a model that effectively leverages temporal information using \textit{streaming} strategy in our proposed temporal fusion module to improve the temporal consistency and quality of vectorized local HD map.
    \item We identify and address significant fairness concerns within the current evaluation setting by establishing a fairer benchmark. In both original and new settings, our method consistently outperforms existing state-of-the-art approaches.
\end{itemize}
\begin{figure*}[ht]
  \centering
   \includegraphics[width=0.9\linewidth]{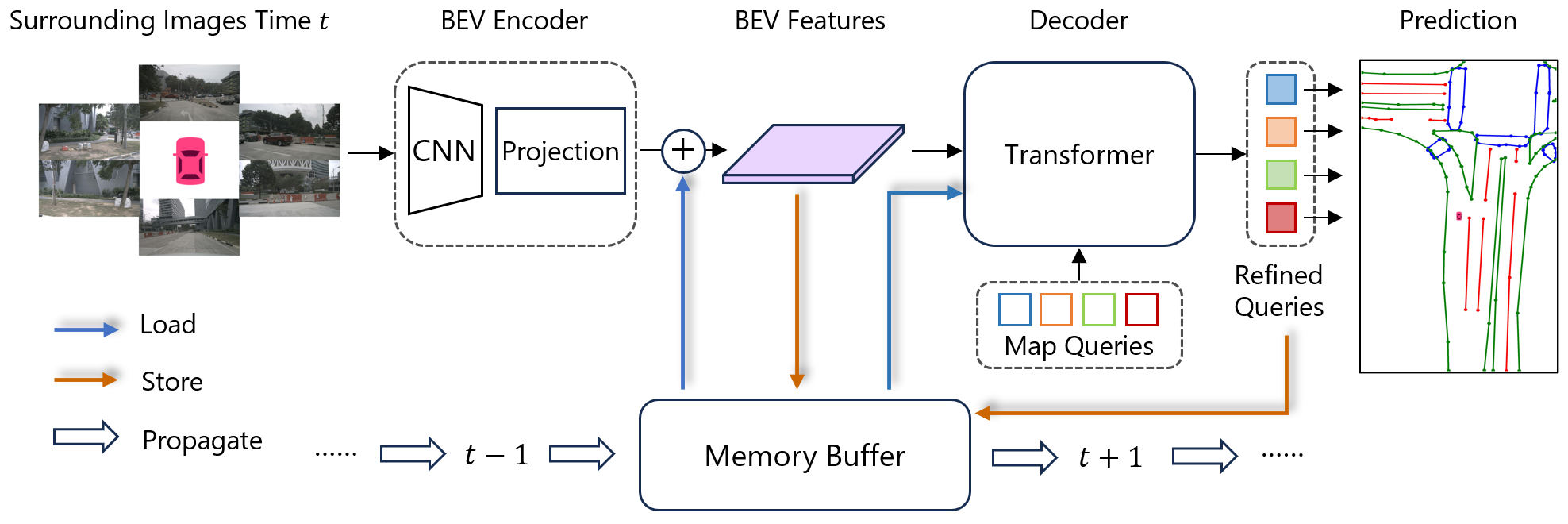}
   \caption{Pipeline of proposed model. Our model architecture comprises three main components: a general image backbone equipped with a BEV encoder for BEV feature extraction, a transformer decoder utilizing Multi-Point Attention for generating predictions, and a memory buffer to store propagated memory features.}
   \label{fig:pipeline}
\end{figure*}
%------------------------------------------------------------------------
\section{Related Works}
\label{sec:related}

%-------------------------------------------------------------------------
\subsection{Online Vectorized Local HD Map Construction}
In recent times, there has been a significant focus on utilizing onboard sensors in autonomous driving vehicles for the construction of vectorized local HD maps. HDMapNet~\cite{li2021hdmapnet} initially generates Bird's-Eye-View (BEV) semantic segmentations, followed by a heuristic and time-consuming pose-processing step to generate vectorized map instances. They also propose using mean Average Precision (mAP) as an evaluation metric.
VectorMapNet~\cite{liu2023vectormapnet} introduces the first end-to-end model that utilizes transformers. It employs a DETR~\cite{carion2020endtoend} decoder to detect map elements and subsequently refines them with an auto-regressive transformer, enabling the construction of fine-grained shapes. However, the auto-regressive model necessitates a long training schedule and leads to reduced inference speed.
MapTR~\cite{MapTR} adopts a one-stage transformer approach to decode map elements using hierarchical queries. Nevertheless, its performance suffers when extending to a wider perception range due to the complex associations among numerous queries.
BeMapNet\cite{qiao2023endtoend} utilizes B'ezier curves along with hand-crafted rules to model map elements.

\subsection{Bird's-Eye-View Perception}
BEV perception techniques have been extensively studied in the domains of 3D object detection and BEV segmentation tasks. Lift-Splat-Shoot~\cite{philion2020lift} proposes using per-pixel predicted depth to lift image features to 3D space. BEVFormer\cite{li2022bevformer} employs deformable attention operations to aggregate image features using learnable BEV queries. SimpleBEV~\cite{harley2022simple} utilizes a variant of Inverse Perspective Mapping~\cite{mallot1991inverse} (IPM) to sample features from 2D images to predefined BEV anchor points.

\subsection{Temporal Modeling for Camera-Based 3D Object Detection}
Inferring 3D space directly from a single-frame camera image is inherently challenging. Recent advancements in camera-based 3D object detection have explored leveraging temporal information to enhance perception outcomes. Some approaches (BEVDet4D~\cite{huang2022bevdet4d}, BEVFormer v2~\cite{yang2022bevformer}) employ a \textit{stacking} strategy, where multiple historical frames or features are stacked and processed together in a single forward pass. However, this strategy incurs significant computational and memory costs that scale linearly with the number of \textit{stacked} frames, thereby reducing training and inference speed while consuming substantial GPU memory. Consequently, the number of \textit{stacked} frames is often limited, resulting in only short-term temporal fusion.
In contrast, recent methods including VideoBEV~\cite{han2023exploring}, StreamPETR~\cite{wang2023exploring} and Sparse4D v2~\cite{lin2023sparse4d} introduce a \textit{streaming} fusion strategy. This strategy treats image sequences as streaming data and processes each frame individually, utilizing memory features propagated from the previous frame. Compared to the \textit{stacking} strategy, the \textit{streaming} strategy enables longer temporal associations while saving GPU memory and reducing latency. VideoBEV~\cite{han2023exploring} propagates BEV features as memory features. Sparse4D v2~\cite{lin2023sparse4d}, as our concurrent work, propagates object queries as memory features. 

\section{StreamMapNet Model}
\subsection{Overall Architecture}
Our model processes sequences of synchronized multi-view images, collected by autonomous vehicles, to create local HD maps. These maps are represented as a set of vectorized instances, each instance consisting of a class label and a polyline parameterized by a sequence of points $\boldsymbol P=\left\{{(x_i, y_i)}\right\}_{i=1}^{N_p}$.

As demonstrate in Figure \ref{fig:pipeline}, our model architecture comprises three main components: a general image backbone equipped with a Bird's Eye View (BEV) encoder for BEV feature extraction, a transformer decoder utilizing Multi-Point Attention for generating predictions, and a memory buffer to store propagated memory features.

\subsection{BEV Feature Encoder}
A shared CNN image backbone is first employed to extract 2D features from multi-view images. Subsequently, these features are aggregated and processed by a Feature Pyramid Network~\cite{lin2017feature} (FPN). Finally, a BEV feature extractor is applied to lift 2D features to BEV space to obtain the BEV feature $\mathcal{F}_\mathrm{BEV} \in \mathbb{R}^{C\times H\times W}$.

\subsection{Decoder Transformer}
While the DETR~\cite{carion2020endtoend} Transformer decoder has demonstrated potency in 3D object detection models operating on 2D BEV features (BEVFormer\cite{li2022bevformer}), its application to HD map construction is not straightforward due to fundamental differences between the tasks. Our approach comprises two key elements in the design of our decoder.

% \subsubsection{Query Design}
% Existing solutions, such as the MapTR, use hierarchical queries (point queries and instance queries) to associate with points on the polylines of map elements, leading to approximately 1,000 point queries in total. However, this design can impede the model's ability to establish associations among a large number of point queries when extending to wider perception range.
\noindent\textbf{Query Design.} Our approach assigns one query to each map instance, which could be a complete pedestrian crossing or a continuous road boundary, resulting in total $N_q$ queries. Conceptually, each query encodes both semantic and geometric information of a map element instance, thus enhancing global scene understanding during self-attention operations. When doing bipartite matching in training, each query matches a ground-truth instance or a background class. We leave the matching cost part to section \ref{sec:loss}.
During the decoding stage, each query generates a class score and $N_p$ point coordinates through Multi-Layer Perceptrons~(MLPs).

\begin{figure}[t]
  \centering
   \includegraphics[width=1.0\linewidth]{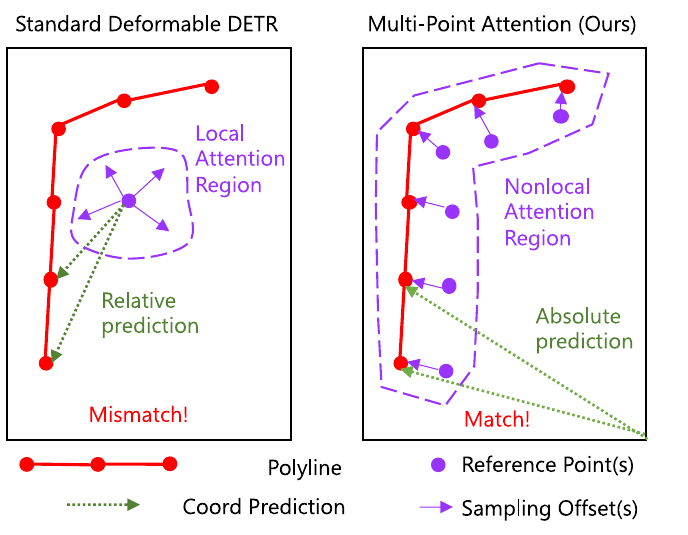}
   \caption{This illustration contrasts conventional deformable DETR with \textbf{Multi-Point Attention}. The former one restrict attention to a localized area, which mismatches the elongate shapes of map elements. Our solution builds a more flexible, non-local attention region.
   }
   \label{fig:MPA}
   \vspace{-1em}
\end{figure}
\noindent\textbf{Multi-Point Attention.} To fit with our query design, we replace the conventional deformable DETR~\cite{zhu2020deformable} design with our proposed \textbf{Multi-Point Attention} in cross attention operation, as demonstrated in Figure \ref{fig:MPA}.
In 3D object detection, an object is typically \textit{local}, occupying a small area close to its center in BEV space. The conventional deformable DETR assigns a reference point to each query as an anchor for collecting features from the constructed BEV features. At every transformer layer, this reference point is adjusted towards the object's center by predicting a residual offset relative to its previous location. For the sake of brevity, we present the formulation of the $i$-th layer below, omitting the self-attention and feed-forward network components:
\begin{gather}
    \boldsymbol{O}_i = \mathrm{Offset\_Embed}(Q_{i-1}) \\
    \boldsymbol{W}_i = \mathrm{Weight\_Embed}(Q_{i-1})\\
    Q_{i} = \sum_{j=1}^{N_\mathrm{off}}\boldsymbol{W}_i^j\cdot
    \mathrm{DA}(Q_{i-1}, R_i+\boldsymbol{O}_i^j, \mathcal{F}_{\mathrm{BEV}})\label{eq:da} \\
    R_{i+1} = \mathrm{sigmoid} (\mathrm{sigmoid}^{-1}(R_i) + \mathrm{Reg}_i(Q_i))
    \label{eq:eq_predrefine}
\end{gather}
Here, $\mathrm{DA}(Q, x, \mathcal{F})$ denotes the deformable attention operation that uses $Q$ as a query to collect features at location $x$ on $\mathcal{F}$. $\boldsymbol{O}_i$ represents the sampling offsets, $N_\mathrm{off}$ the number of sampling offsets for each query, $\boldsymbol{W}_i$ the sampling weights, $R_i$ the reference points, and $\mathrm{Reg}_i$ the object center regression branch. The subscript $i$ indicates the $i$-th layer and supercript $j$ indicates the $j$-th element.

However, a map element may display a highly irregular and elongated shape, making it \textit{nonlocal} in BEV space. Therefore, our method use the $N_p$ predicted points, rather than the object center, from the previous layer as the reference points in the current layer. While facilitating long-range attention in BEV space, this approach maintains low complexity: $O(N_p)$, compared to $O(HW)$ for global attention. We employ a shared MLP for all layers as the regression branch to predict the absolute coordinates rather than a residual offset. The $i$-th layer can then be formulated as follows:
\begin{gather}
    \boldsymbol{O}_i = \mathrm{Offset\_Embed}(Q_{i-1}) \\
    \boldsymbol{W}_i = \mathrm{Weight\_Embed}(Q_{i-1})\\
    % \begin{aligned}
    % Q_{i} =& \sum_{j=1}^{N_p}\sum_{k=1}^{N_\mathrm{off}}
    % \boldsymbol{W}_i^{(j-1)\cdot N_\mathrm{off}+k}\cdot\\
    % &\mathrm{DA}(Q_{i-1}, \boldsymbol P_i^j+\boldsymbol{O}_i^{(j-1)\cdot 
    % N_\mathrm{off}+k}, \mathcal{F}_{\mathrm{BEV}}) \label{eq:eq_mpa}
    % \end{aligned}
    \scriptsize{Q_{i} = \sum_{j=1}^{N_p}\sum_{k=1}^{N_\mathrm{off}}
    \boldsymbol{W}_i^{(j-1)\cdot N_\mathrm{off}+k}\cdot \mathrm{DA}(Q_{i-1}, \boldsymbol P_i^j+\boldsymbol{O}_i^{(j-1)\cdot 
    N_\mathrm{off}+k}, \mathcal{F}_{\mathrm{BEV}})} \label{eq:eq_mpa} \\
    \boldsymbol P_{i+1} =  \mathrm{sigmoid}(\mathrm{Reg}(Q_i))
    \label{eq:eq8}
\end{gather}
Please note that in this context, $\boldsymbol P^j_i$ represent the coordinates of the $j$-th points on the predicted polyline at $i$-th layer.

\subsection{Temporal Fusion}
This section describes two temporal fusion modules that integrate temporal information from memory features into the current frame: \textbf{Query Propagation} and \textbf{BEV Fusion}.

\begin{figure}[t]
  \centering
   \includegraphics[width=1.0\linewidth]{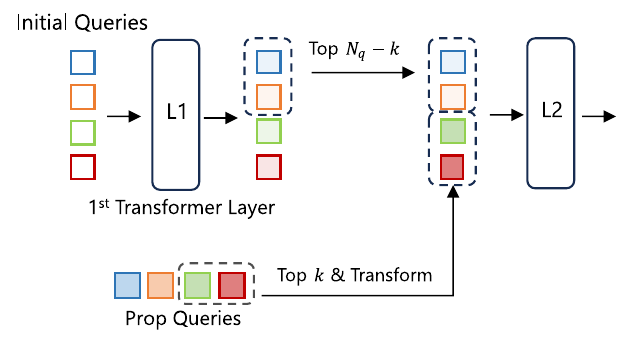}
   \caption{The top $k$ refined queries are propagated from the previous frame. Post transformation, these queries are integrated with the top $N_q - k$ queries from the first Transformer layer, assembling a refreshed set of $N_q$ queries.}
   \label{fig:QueryProp}
   \vspace{-1ex}
\end{figure}

\noindent\textbf{Query Propagation.} In map construction scenarios, all map instances are static, suggesting that instances in the current frame are likely to persist in subsequent frames. This motivates us to propagate queries with the highest $k$ confidence scores to the next frame, providing a valuable positional prior and retaining temporal features across all historical frames. Given that we employ ego-coordinates, these propagated queries must be transformed before utilization. We employ a MLP with a residual connection to facilitate this transformation in the latent space.
\begin{equation}
    Q_{t} = \phi_\mathrm{t}\left(
    \mathrm{Concat}(Q_{t-1}, \mathrm{flatten}(\boldsymbol{T}))
    \right) + Q_{t-1}
\end{equation}
Here, $\boldsymbol{T}$ denotes a standard $4\times 4$ transformation matrix between the coordinate systems of two frames. We also convert the predicted $N_p$-point polyline to the new coordinate system to serve as the initial reference points for the propagated queries.
\begin{equation}
    \boldsymbol{P}_{t} = \boldsymbol T\cdot \mathrm{homogeneous}(\boldsymbol{P}_{t-1})_{:,0:2}
\end{equation}
Figure \ref{fig:QueryProp} illustrates the incorporation of propagated queries into the decoder. We use initial object queries in the first decoder. Post the initial decoder layer, we select the top $N_q - k$ queries based on confidence scores as potential foreground queries, and concatenate them with propagated queries. This approach aligns with the principles of Sparse4D v2~\cite{lin2023sparse4d}. We add an auxiliary transformation loss to assist transformation learning:
\begin{gather}
    \hat{\boldsymbol{P}} = \mathrm{Reg}(Q_t)\\
    \mathcal{L}_\mathrm{trans} = \sum_{j=1}^{N_p} 
    \mathcal{L}_\mathrm{SmoothL1}(\hat{\boldsymbol{P}}^j, \boldsymbol{P}^j_t)
    \label{eq:auxloss}
\end{gather}

\begin{figure}[t]
  \centering
   \includegraphics[width=1.0\linewidth]{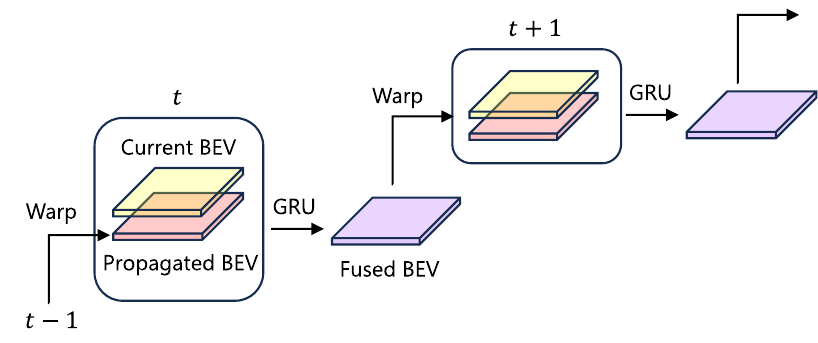}
   \caption{The propagated BEV feature, serving as a recurrent memory, is warped and updated for each frame.}
   \label{fig:BEVFusion}
   \vspace{-1ex}
\end{figure}

\noindent\textbf{BEV Fusion.} While \textbf{Query Propagation} operates temporal association on sparse queries, the dense BEV features can also benefit from incorporating historical features. We recurrently propagate BEV features and warp them based on the ego vehicle's pose, as illustrated in Figure \ref{fig:BEVFusion}. Drawing inspiration from the Neural Map Prior~\cite{xiong2023neuralmapprior}, we employ a Gated Recurrent Unit~\cite{chung2014empirical} (GRU) to fuse these BEV features. To ensure training stability, we introduce a layer normalization operation in the final step.
\begin{gather}
    \Tilde{\mathcal{F}}^{t-1}_\mathrm{BEV} = \mathrm{Warp}\left(\mathcal{F}^{t-1}_\mathrm{BEV}, \boldsymbol{T}\right) \\
    \mathcal{F}^{t}_\mathrm{BEV} = \mathrm{LayerNorm}\left(
    \mathrm{GRU}\left(
    \Tilde{\mathcal{F}}^{t-1}_\mathrm{BEV}, \mathcal{F}^{t}_\mathrm{BEV}
    \right)
    \right)
\end{gather}

% \vspace{1ex}
\subsection{Matching Cost and Training Loss}
\label{sec:loss}
Our model adopts an end-to-end training approach. We employ standard bipartite matching to pair predicted map instances with their ground-truth counterparts, denoted as ${(c_i, \boldsymbol P_i)}{i=1}^{N\mathrm{gt}}$. Predicted instances are represented as ${(\hat c_i, \hat {\boldsymbol P}i)}{i=1}^{N_q}$. In this context, we slightly modify the notation $\hat{\boldsymbol{P}}_i$ to indicate the predicted polyline of the $i$-th query, diverging from equation \ref{eq:eq8}.
The polyline-wise matching cost is defined as follows:
\begin{equation}
    \mathcal{L}_\mathrm{line}(\hat{\boldsymbol{P}}, \boldsymbol{P}) = 
    \min_{\gamma \in \Gamma}
    \frac{1}{N_p}\sum_{j=1}^{N_p} 
    \mathcal{L}_\mathrm{SmoothL1}(\hat{p}_j, p_{\gamma(j)})
\end{equation}
In this case, $\hat{p}_j$ signifies the $j$-th point of $\hat{\boldsymbol{P}}$. The permutation group introduced by MapTR~\cite{MapTR} is denoted by $\Gamma$. For the classification matching cost, we utilize Focal Loss. The final matching cost is then expressed as:
\begin{equation}
    \begin{aligned}
        \mathcal{L}_\mathrm{match}\left(
        (\hat c_i, \hat {\boldsymbol P}_i), (c_i, \boldsymbol P_i)
        \right)
        = \\
        \lambda_1\mathcal{L}_\mathrm{line}(\hat{\boldsymbol{P}}, \boldsymbol{P})
        +
        \lambda_2\ \mathcal{L}_\mathrm{Focal}(\hat{c}_i, c_i)
    \end{aligned}
\end{equation}
Despite the auxiliary loss presented in equation \ref{eq:auxloss}, the training loss mirrors the structure of the matching cost and is defined as:
\begin{equation}
    \mathcal{L}_\mathrm{train}=
    \lambda_1\mathcal{L}_\mathrm{line}
    +
    \lambda_2\ \mathcal{L}_\mathrm{Focal}
    +\lambda_3 \mathcal{L}_\mathrm{trans}
\end{equation}

\section{Experiments}
\begin{table*}[t]
  \centering
  \resizebox{0.95\textwidth}{!}{
    \begin{tabular}{c|lccc|ccccc}
    \toprule
    Range & Method & Backbone & Image Size & Epoch & AP$_{ped}$ & AP$_{div}$ & AP$_{bound}$ & mAP & FPS\\
    \midrule
    \multirow{3}{*}{$60\times 30\,m$} &  
    VectorMapNet & R50 & $384\times 384$ & 120 & 35.6 & 34.9 & 37.8 & 36.1 & 5.5 \\
    \multirow{3}{*}{ } & MapTR & R50 & $608\times 608$ & 30 & 48.1 & 50.4 & 55.0 & 51.1& \textbf{18.0}  \\
    \multirow{3}{*}{ } &StreamMapNet (Ours) & R50 & $608\times 608$ & 30 & \textbf{56.9} & \textbf{55.9} & \textbf{61.4} & \textbf{58.1} & 14.2 \\
    \midrule
    \multirow{3}{*}{$100\times 50\,m$} & VectorMapNet & R50 & $384\times 384$ & 120 & 32.4 & 20.6 & 24.3 & 25.7 & 5.5  \\
    \multirow{3}{*}{} & MapTR & R50 & $608\times 608$ & 30 & 46.3  & 36.3  & 38.0  & 40.2  &  \textbf{18.0}  \\
    \multirow{3}{*}{} & {StreamMapNet (Ours)} & R50 & $608\times 608$ & 30 & \textbf{60.5} & \textbf{44.4} & \textbf{48.6} & \textbf{51.2} &  14.2 \\
    
    \bottomrule
  \end{tabular}
  }
  \caption{Performance comparison of various methods on the new Argoverse2 split at both $30m$ and $50m$ perception ranges. StreamMapNet notably outperforms other methods in all categories, exhibiting robustness to long perception ranges due to its integration of temporal association and long-range attention mechanism.}
  \label{tab:av2_100x50}
  \vspace{-1ex}
\end{table*}
\subsection{Rethinking on Datasets}
\label{sec:rethink}
NuScenes~\cite{caesar2020nuscenes}, a popular benchmark in autonomous driving research, offers around 1000 scenes with six synchronized cameras and precise ego-vehicle poses. Most online HD map construction models test their performance on this dataset, following the official 700/150/150 split for training/validation/testing scenes. However, this split, intended for object detection tasks, falls short of map construction.

We found an overlap of over $84\%$ of locations between the training and validation sets. This overlap might not be problematic for object detection, given the significant variance in objects across different traversals. Yet, the map remains essentially unchanged, which means a model could simply memorize location-HD map pairs from the training set and perform exceptionally well on the validation set. However, such a model would completely fail to generalize to new scenes. This clearly contradicts the essence of online map construction: the goal is to develop models that can generalize to unseen environments and adapt to potential map changes, rather than simply memorizing the training set.

Similar issues were detected with Argoverse2~\cite{Argoverse2}, another dataset with 1000 scenes from six cities, where we identified a $54\%$ overlap between validation and training locations. To address this, we use new training/validation splits for both datasets that minimize overlap and ensure balanced location, object, and weather conditions distribution. We employ Roddick and Cipolla's \cite{roddick2020predicting} splits for NuScenes, and introduce a new split for Argoverse2. Both splits result in a 700/150 division for training/validation scenes. It will be released along with the code. It's worth noting that when we discuss results on a specific split, we are referring to models trained on that split's training set and evaluated on its validation set. 
While we primarily compare performance on these new splits, we also present results on the original splits for thorough and equitable comparison with existing methods. 

\subsection{Implementation Details}
Our model trains on $8$ GTX3090 GPUs with a batch size of $32$, using an AdamW optimizer \cite{loshchilov2018fixing} with a learning rate of $5\times 10^{-4}$. We adopt ResNet50 \cite{he2016deep} as backbones and use BEVFormer~\cite{li2022bevformer} with a single encoder layer for BEV feature extraction, consistent with MapTR~\cite{MapTR}. The model trains for $24$ epochs on the NuScenes dataset and $30$ epochs on Argoverse2. We set $N_q=100$, $N_\mathrm{off}=1$, $N_p=20$, $k=33$, $\lambda_1=50.0$, $\lambda_2=5.0$, $\lambda_1=5.0$ as the hyperparameters for all settings and perception ranges without further tuning. 

\noindent \textbf{Streaming Training.} We adopt the \textit{streaming} training strategy for temporal fusion, as illustrated in Figure \ref{fig:streaming}. Gradients on memory features do not propagate back to previous frames. For each training sequence, we randomly divide it into $2$ splits at the start of each training epoch to foster more diverse data sequences. During inference, we use the entire sequences. To stabilize streaming training, we train the initial $4$ epochs with single-frame input, inspired by SOLOFusion~\cite{Park2022TimeWT}.

\subsection{Metrics}
In line with existing works, we consider three types of map elements: pedestrian crossings, lane dividers, and road boundaries. We enlarge the perception range to cover an area of $50\,m$ front and back, and $25\,m$ left and right, aligning with the scope of 3D object detection tasks. Concurrently, we also present results for a smaller range ($30\,m$ front and back, $15\,m$ left and right), as used by prior works.
We adopt Average Precision (AP) as the evaluation metric proposed in \cite{li2021hdmapnet} and \cite{liu2023vectormapnet}. AP calculations are conducted under distinct thresholds: $\{1.0\,m, 1.5\,m, 2.0\,m\}$ for $50\,m$ range, and $\{0.5\,m, 1.0\,m, 1.5\,m\}$ for the $30\,m$ range.

\subsection{Comparison with Baselines}
We implemented VectorMapNet\cite{liu2023vectormapnet} and MapTR~\cite{MapTR} using their official codebases, altering only the perception range and training-validation split. VectorMapNet's input image size was adjusted to suit the memory of an RTX3090 GPU. For MapTR, BEVFormer~\cite{li2022bevformer} was used as the BEV feature extractor to ensure a fair comparison with our model. As BeMapNet's~\cite{qiao2023endtoend} code is not publicly available, we could only compare with their reported results on the original NuScenes split with a $30\,m$ perception range.

\vspace{-1em}
\subsubsection{Performance on Argoverse2 Dataset}
\quad Argoverse2 dataset originally provide $10\,$Hz camera frame rate. To align with the NuScenes setup, we set the camera frame rate to $2\,$HZ.

\noindent\textbf{New Split.} Table \ref{tab:av2_100x50} showcases the performance comparison on the new Argoverse2 split. We report results for both $30\,m$ and $50\,m$ perception ranges. At both both perception ranges, StreamMapNet demonstrates superior performance over other methods across all categories while maintaining a online inference speed, showing the effectiveness of our approach. Existing methods experience a significant drop in mAP when the perception range is increased.In contrast, our method is more robust due to the incorporation of temporal associations and long-range attention mechanism.
\begin{table*}[h]
  \centering
  \resizebox{0.95\textwidth}{!}{
    \begin{tabular}{c|lccc|ccccc}
    \toprule
    Range & Method & Backbone & Image Size & Epoch & AP$_{ped}$ & AP$_{div}$ & AP$_{bound}$ & mAP & FPS\\
    \midrule
    \multirow{3}{*}{$60\times 30\,m$} &  
    VectorMapNet & R50 & $256\times 480$ & 120 & 15.8 & 17.0 & 21.2 & 18.0 & 3.8  \\
    \multirow{3}{*}{ } & MapTR & R50 & $480\times 800$ & 24 & 6.4 & 20.7 & 35.5 & 20.9 &  \textbf{16.0} \\
    \multirow{3}{*}{ } &StreamMapNet (Ours) & R50 & $480\times 800$ & 24 & \textbf{29.6} & \textbf{30.1} & \textbf{41.9} & \textbf{33.9} & 13.2 \\
    \midrule
    \multirow{3}{*}{$100\times 50\,m$} & VectorMapNet & R50 & $256\times 480$ & 120 & 12.0 & 8.1 & 6.3 & 8.8 & 3.8  \\
    \multirow{3}{*}{} & MapTR & R50 & $480\times 800$ & 24 & 8.3 & 16.0 & 20.0 & 14.8 & \textbf{16.0}  \\
    \multirow{3}{*}{} & StreamMapNet (Ours) & R50 & $480\times 800$ & 24 & \textbf{24.8} & \textbf{19.6} & \textbf{24.7} & \textbf{23.0} & 13.2 \\
    \bottomrule
  \end{tabular}
  }
  \caption{Performance comparison with baseline methods on the new NuScenes split at both $30\,m$ and $50\,m$ perception ranges. SteamMapNet outperforms existing methods. While StreamMapNet exhibits superior performance compared to existing methods, all approaches experience a performance reduction relative to the results obtained using the Argoverse2 new split.}
  \label{tab:nusc_new}
\end{table*}
\begin{table}[h]
  \centering
  \resizebox{0.45\textwidth}{!} {
      \begin{tabular}{lcc|c}
        \toprule
        Method & Image Size & Epoch & mAP\\
        \midrule
        VectorMapNet & $384\times 384$ & 120 & 30.2 \\
        MapTR & $608\times 608$ & 30 & 47.5 \\
        StreamMapNet (Ours) & $608\times 608$ & 30 & \textbf{57.7} \\
        \bottomrule
      \end{tabular}
    }
  \caption{Performance comparison on the original Argoverse2 training/validation split at a $50\,m$ perception range. Our method consistently outperforms other methods.}
  \vspace{-1em}
  \label{tab:av2_old_100x50}
\end{table}

\noindent\textbf{Original Split.} Table \ref{tab:av2_old_100x50} presents performance results on the original training/validation split at the $50\,m$ range for a comprehensive comparison. StreamMapNet consistently surpasses other methods on the original split by a significant margin of at least $10.2$ mAP. A significant performance gap can be found when comparing results between the new split and the original split (Table \ref{tab:av2_100x50}), indicating the overfitting problem of the original split cannot be ignored.
\vspace{-1em}
\subsubsection{Performance on NuScenes Dataset}
\noindent\textbf{New Split.} Table \ref{tab:nusc_new} compares performance on the new NuScenes split at both $30\,m$ and $50\,m$ perception ranges. Our method shows a considerable improvement, surpassing existing methods by $13.0$ mAP at the $30\,m$ range and $8.2$ mAP at the $50\,m$ range.

A decline in performance is observed across all methods when compared to results on the Argoverse2 dataset (Table \ref{tab:av2_100x50}). We attribute this to two main reasons: (1) The Argoverse2 dataset offers images from more cameras with higher resolutions ($7$ cameras with resolution $1550\times 2048$), whereas NuScenes provides images from $6$ cameras with a resolution of $900\times 1600$. Cameras in Argoverse2 are positioned at higher viewpoints, thus providing a longer viewing range. (2) The Argoverse2 dataset contains more diverse training data with locations across six different cities, in contrast to NuScenes' two cities. Despite these challenges, NuScenes remains a valuable benchmark for evaluating online map construction tasks.

\begin{table}[h]
  \centering
  \resizebox{0.45\textwidth}{!} {
  \begin{tabular}{lcc|c}
    \toprule
    Method & Image Size & Epoch & mAP\\
    \midrule
    VectorMapNet & $256\times 480$ & 110 & 40.9 \\
    MapTR & $480\times 800$ & 24 & 48.7 \\
    BeMapNet~\cite{qiao2023endtoend} & $512\times 896$ & 30 & 59.8\\
    StreamMapNet (Ours) & $480\times 800$ & 24 & \textbf{62.9} \\
    
    \bottomrule
  \end{tabular}
  }
  \caption{Performance comparison on the original NuScenes split with $30\,m$ range, a widely used benchmark for evaluating online map construction tasks. StreamMapNet outperforms existing methods.
  However, this validation set in this split is prone to overfitting.}
  \label{tab:nusc_old}
  \vspace{-1em}
\end{table}
\noindent\textbf{Original Split.} A majority of existing methods primarily evaluate their results on the original NuScenes split at a perception range of $30\,m$. Despite the tendency for overfitting within this setting, we provide a comparison of StreamMapNet's performance against these methods to ensure comprehensive analysis. As seen in Table \ref{tab:nusc_old}, StreamMapNet outperforms existing methodologies even with fewer or equivalent training epochs. A comparison with the results in Table \ref{tab:nusc_new} reveals that transitioning to the new split induces approximately a $50\%$ performance decrease across all methods, which fully substantiates the concerns raised in section \ref{sec:rethink}. The original split's validation set seems prone to overfitting, rendering it less reliable when evaluating the generalization capabilities of online map construction models, especially those utilizing a large backbone with a higher capacity for memorization.

\begin{figure*}[t]
  \centering
   \includegraphics[width=1.0\linewidth]{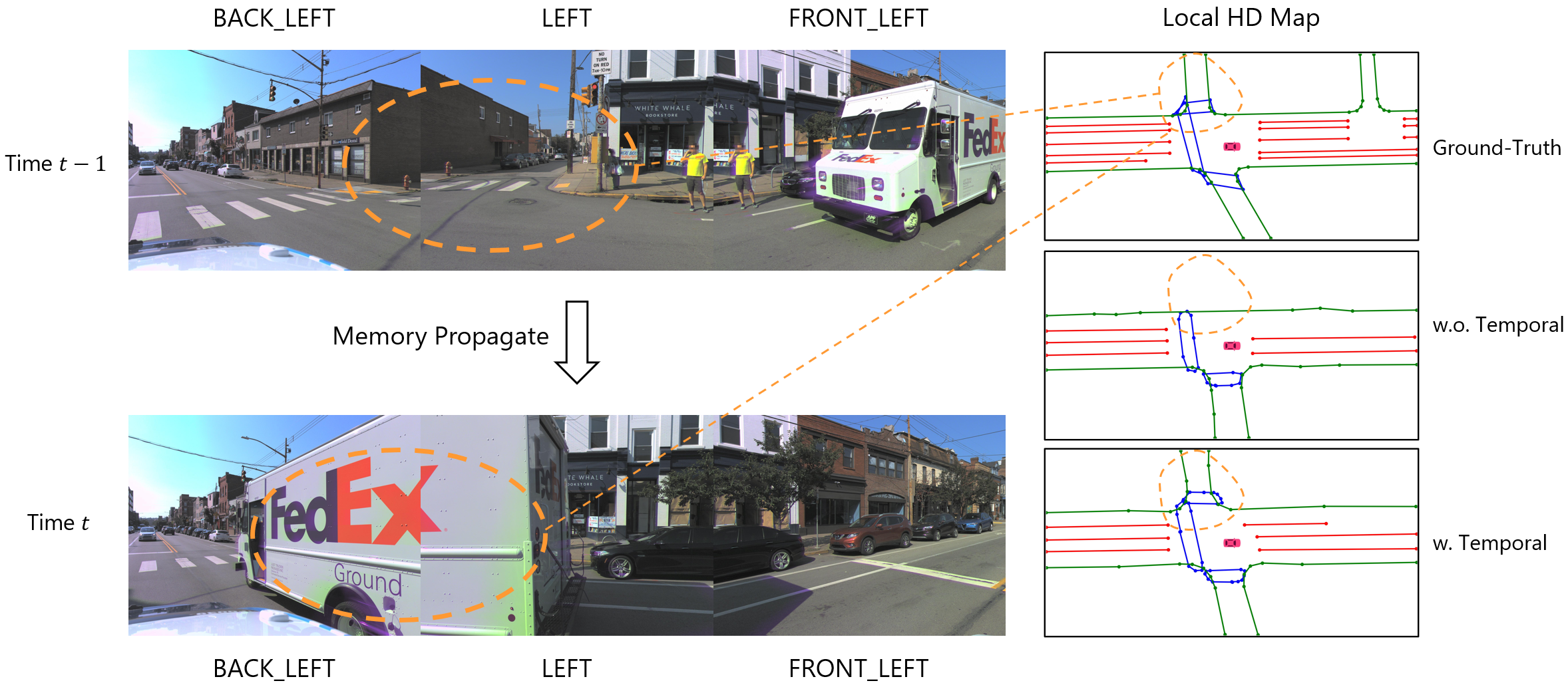}
   \caption{We compare models with and without temporal information in occlusion scenarios. At Time $t$, the crossroad is occluded by a white truck (highlighted by the orange circle). The model incorporating temporal information successfully constructs the road structure, while the single-frame model falls short. In the HD maps, \textit{green} lines denote road boundaries, \textit{red} lines indicate lane denote, and \textit{blue} lines denote pedestrian crossings.}
   \label{fig:visualize}
   \vspace{-2ex}
\end{figure*}
\subsection{Ablation Studies}
\begin{table}[h]
  \centering
  \resizebox{0.45\textwidth}{!}{
  \begin{tabular}{c|l|c}
    \toprule
    Index & Method & mAP\\
    \midrule
    (a)&Single-frame baseline w. relative predict& 33.7\\
    (b)&$-$ Multi-Point attention & - \\
    (c)&$+$ Direct predict & 41.7 \\
    (d)&$+$ Query propagation (w.o. trans. loss) & 42.8 \\
    (e)&$+$ Transformation loss & 43.7 \\
    (f)&$+$ BEV fusion & 46.1 \\
    (g)&$+$ Image size $608\times 608$ & 51.2 \\
    \midrule
    & StreamMapNet & \textbf{51.2}\\   
    \bottomrule
  \end{tabular}
  }
  \caption{Ablation study of each component. Starting from a single-frame baseline model to the full model. Each modification contributes to the performance gain.}
  \label{tab:ablation}
  \vspace{-1ex}
\end{table}
We examine the efficacy of each StreamMapNet component through ablation studies, utilizing the new Argoverse2 split at a $50\,m$ perception range. Unless stated otherwise, we adjust image sizes to $384\times 384$. The influence of each component is demonstrated in Table \ref{tab:ablation}.
Initially, we build a single-frame baseline model employing only Multi-Point attention. The term \textit{relative predict} refers to the operation in Equation \ref{eq:eq_predrefine}, a technique commonly found in object detection models. To assess the importance of Multi-Point Attention, we replace it with the conventional deformable design, in which each query is allocated a reference point at its center of gravity (substituting Equation \ref{eq:eq_mpa} with Equation \ref{eq:da}). This replacement hampers model convergence due to the restrictive attention range, validating the necessity of Multi-Point Attention. For integration with Multi-Point Attention, we replace \textit{Predict Refinement} with \textit{Direct Predict}, achieving a robust single-frame model. Progressing from model (d) to (f), we gradually introduce the temporal fusion components, consistently enhancing performance and underscoring the significance of temporal information association in online map construction tasks.

\subsection{Qualitative Analysis}
In this section, we present qualitative results from our StreamMapNet, emphasizing the importance of temporal modeling by comparing it with a single-frame model. Figure \ref{fig:visualize} illustrates a commonplace scenario in autonomous driving where a large truck obstructs part of the camera's field of view. For better visualization, we focus only on images from the left side.
The ground-truth map indicates a crossroad to the left of the ego vehicle, an area briefly hidden by the truck in the current frame. Without the benefit of temporal information, the single-frame model, lacking visual information beyond the truck, fails to accurately replicate the crossroad. However, our model effectively uses temporal information from previous frames to correctly reproduce the road structure. This leads to the generation of a stable and reliable HD map, which is vital to ensure the safety of autonomous vehicles.
\vspace{-1ex}
\section{Conclusion \& Acknowledgement}
In this study, we have proposed an end-to-end model for the online construction of vectorized, local HD maps. By leveraging temporal information, our approach promotes stability in wide-range map perception. Importantly, we scrutinize the prevalent evaluation settings on NuScenes and Argoverse2 datasets and identified improper training/validation division that leads to the overfitting problem. As a remedy, we propose new, non-overlapping splits for both datasets. We hope that these refined splits will foster a more balanced benchmark for future research in this field.

\noindent\textbf{Discussion of Potential Negative Societal Impact.} While our model significantly improves upon existing methods, it may still make false predictions in challenging scenarios, which underlines the necessity for comprehensive safety testing before deploying our model in real autonomous driving vehicles. Moreover, HD map data can be sensitive. The collection and use of such data might violate laws and regulations in certain countries or regions. As such, it is imperative to take the necessary precautions before gathering this kind of data and training our model on it, to ensure the privacy and legal rights of individuals are respected.

\noindent\textbf{Acknowledgement} This work is supported by Tsinghua University Dushi Program.

%------------------------------------------------------------------------
% \section{Final copy}

% You must include your signed IEEE copyright release form when you submit your finished paper.
% We MUST have this form before your paper can be published in the proceedings.

% Please direct any questions to the production editor in charge of these proceedings at the IEEE Computer Society Press:
% \url{https://www.computer.org/about/contact}.

%%%%%%%%% REFERENCES
{\small
\bibliographystyle{ieee_fullname}
\bibliography{egbib}
}

\newpage
\appendix
% \onecolumn

%%%%%%%%% BODY TEXT - ENTER YOUR RESPONSE BELOW
\section{Dataset Statistics}
\begin{table}[h]
  \centering
  \resizebox{0.46\textwidth}{!}{
    \begin{tabular}{@{}c|c|ccc@{}}
    \toprule
    Dataset & Split & Train ($km^2$) & Val ($km^2$) & Overlap ($km^2$) / Ratio \\
    \midrule
    \multirow{2}{*}{Argoverse2} & Original & $3.46$ & $1.04$ & $0.56$ / $54\%$
     \\
    \multirow{2}{*}{ } & New & $3.58$ & $0.54$ & $\mathbf{0.00}$ / $\mathbf{0\%}$ \\
    \midrule
    \multirow{2}{*}{NuScenes} & Original & $2.00$ & $0.92$ & $0.79$ / $85\%$
     \\
    \multirow{2}{*}{ } & New & $1.64$ & $0.54$ & $\mathbf{0.06}$ / $\mathbf{11\%}$ \\
    \bottomrule
  \end{tabular}
  }
  \caption{Summary of the cumulative areas of the training set, validation set, and their overlap in both datasets. The original splits of both datasets exhibit a high overlap ratio, a problem substantially mitigated by our suggested re-division.}
  \label{tab:stat}
\end{table}
In this supplementary material, we provide an in-depth analysis of the NuScenes~\cite{caesar2020nuscenes} and Argoverse2~\cite{Argoverse2} datasets to further emphasize the necessity of re-splitting both datasets for the task of map construction.

Table~\ref{tab:stat} shows the cumulative areas of the training and validation sets and their overlap across different splits of both datasets. We query all locations within a $30,m$ radius circle from the ego-vehicle's position.

In Argoverse2, the initial overlap ratio is $54\%$. By merging the training, validation, and testing sets together, we redivide them into a new train/validation split. We propose that an additional testing set is unnecessary, as all map data is publicly available. A fair testing set cannot be constructed if the ground-truth data is not hidden. Consequently, we divide the entire set of $1,000$ scenes into a $700/150$ train/validation split, ensuring a balanced distribution over objects, weathers, and cities, while completely eliminating any overlaps.

In NuScenes, more than $85\%$ of locations in the validation set appear in the training set in the original split, which raises serious concerns about overfitting. However, the overlap ratio is reduced to $11\%$ after using Roddick and Cipolla's new split~\cite{roddick2020predicting}, substantially mitigating this issue.

\section{Dataset Visualization}
\begin{figure*}[t]
  \centering
   \includegraphics[width=1.0\linewidth]{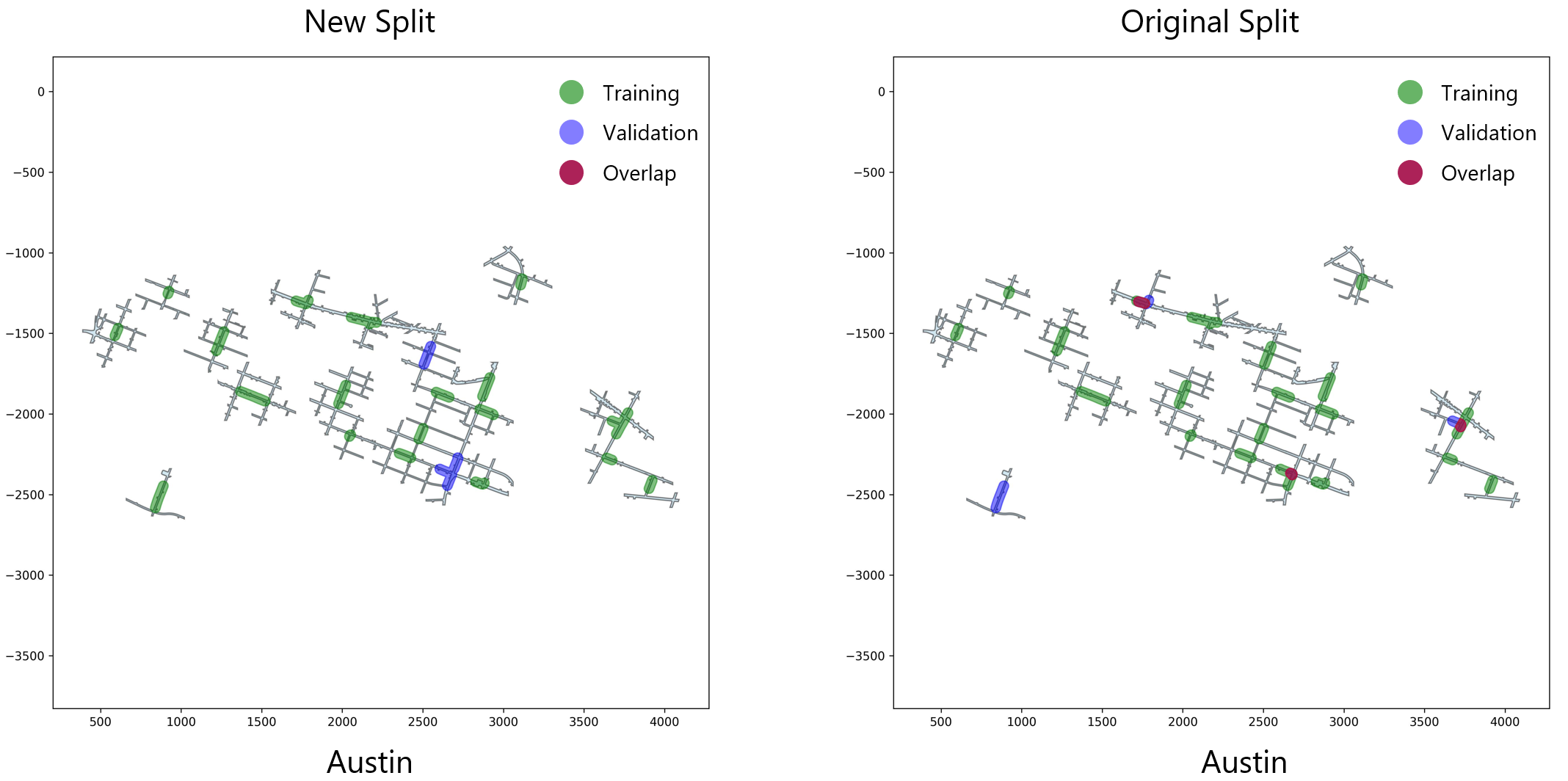}
   \caption{Comparison of the new and original splits in the Austin area of the Argoverse2 dataset. The \textit{green} regions represent the training set, \textit{blue} denotes the validation set, and \textit{red} signifies the overlapping areas. Best viewed in color.}
   \label{fig:ATX}
\end{figure*}

\begin{figure*}[t]
  \centering
   \includegraphics[width=1.0\linewidth]{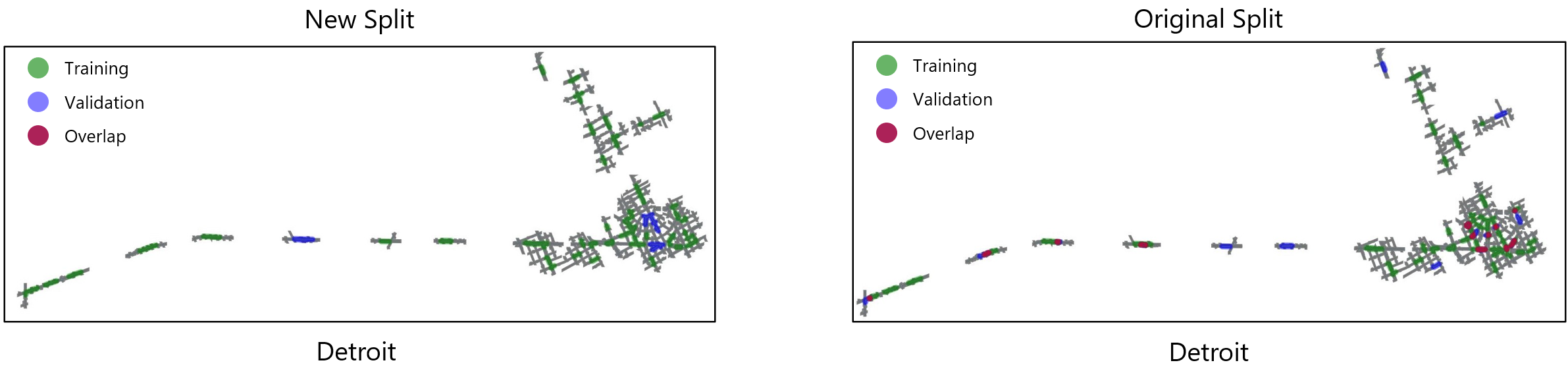}
   \caption{Comparison of the new and original splits in the Detroit area of the Argoverse2 dataset. The \textit{green} regions represent the training set, \textit{blue} denotes the validation set, and \textit{red} signifies the overlapping areas. Best viewed in color.}
   \label{fig:DTW}
\end{figure*}

\begin{figure*}[t]
  \centering
   \includegraphics[width=1.0\linewidth]{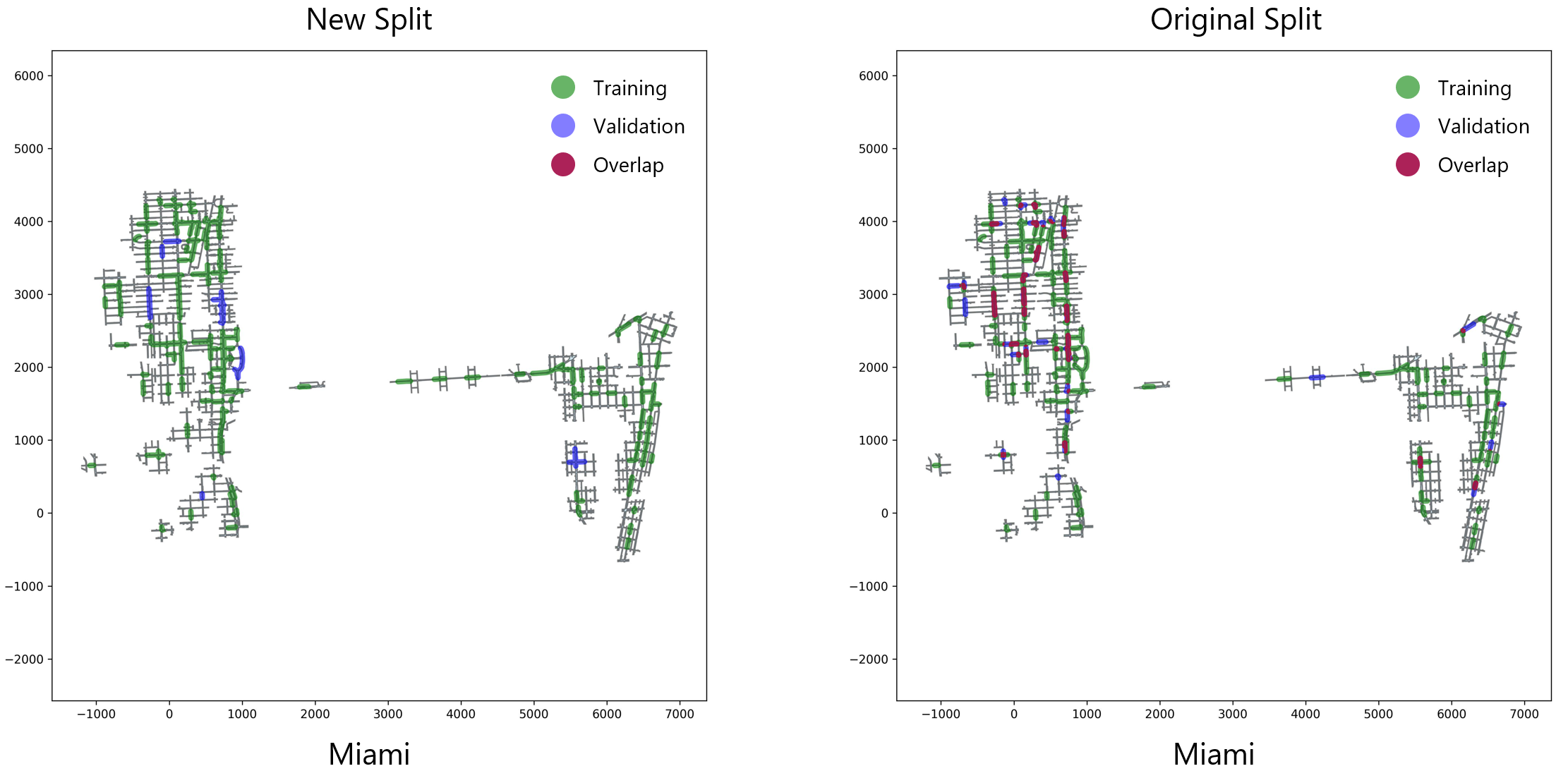}
   \caption{Comparison of the new and original splits in the Miami area of the Argoverse2 dataset. The \textit{green} regions represent the training set, \textit{blue} denotes the validation set, and \textit{red} signifies the overlapping areas. Best viewed in color.}
   \label{fig:MIA}
\end{figure*}

\begin{figure*}[t]
  \centering
   \includegraphics[width=1.0\linewidth]{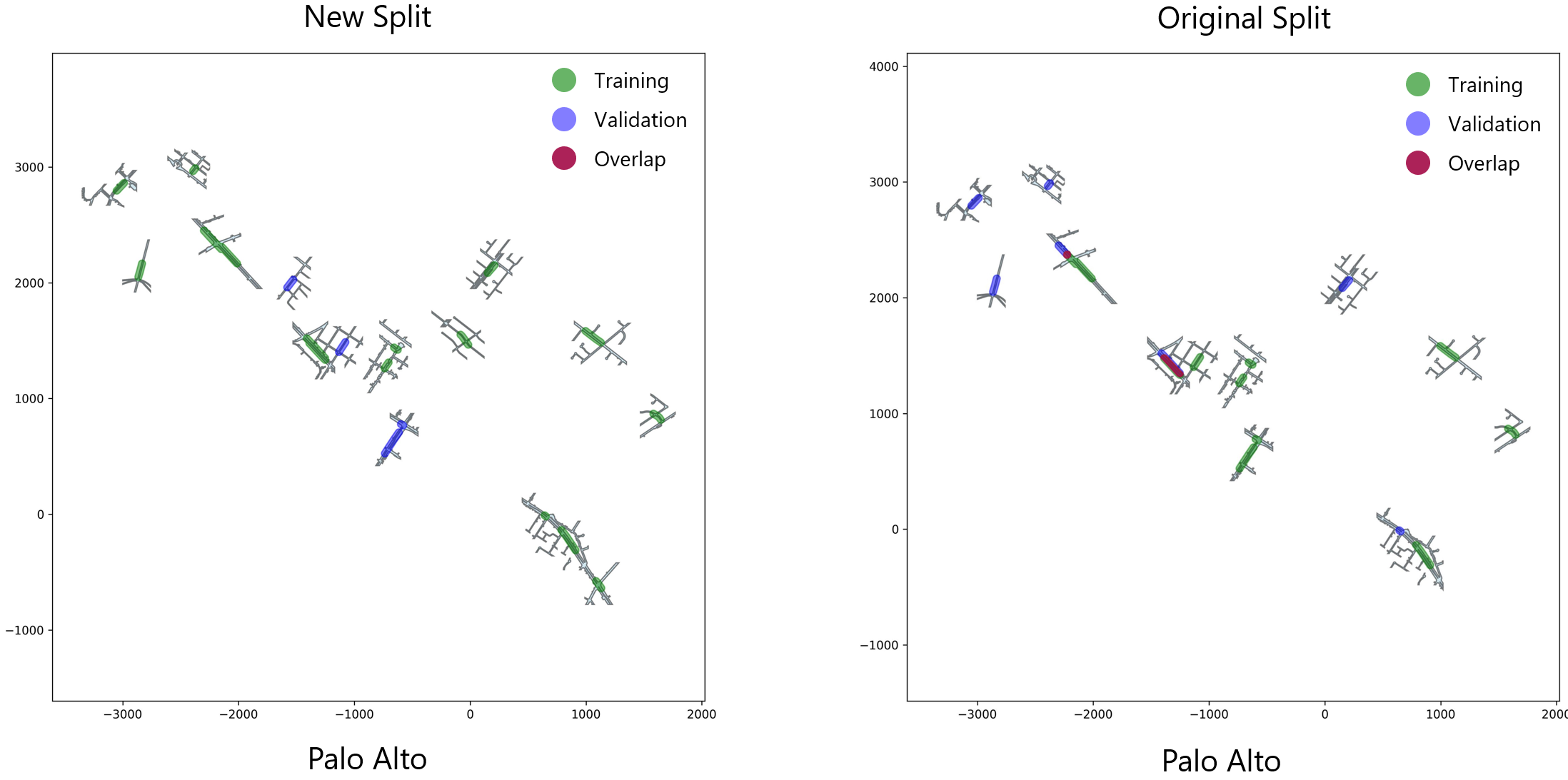}
   \caption{Comparison of the new and original splits in the Palo Alto area of the Argoverse2 dataset. The \textit{green} regions represent the training set, \textit{blue} denotes the validation set, and \textit{red} signifies the overlapping areas. Best viewed in color.}
   \label{fig:PAO}
\end{figure*}

\begin{figure*}[t]
  \centering
   \includegraphics[width=1.0\linewidth]{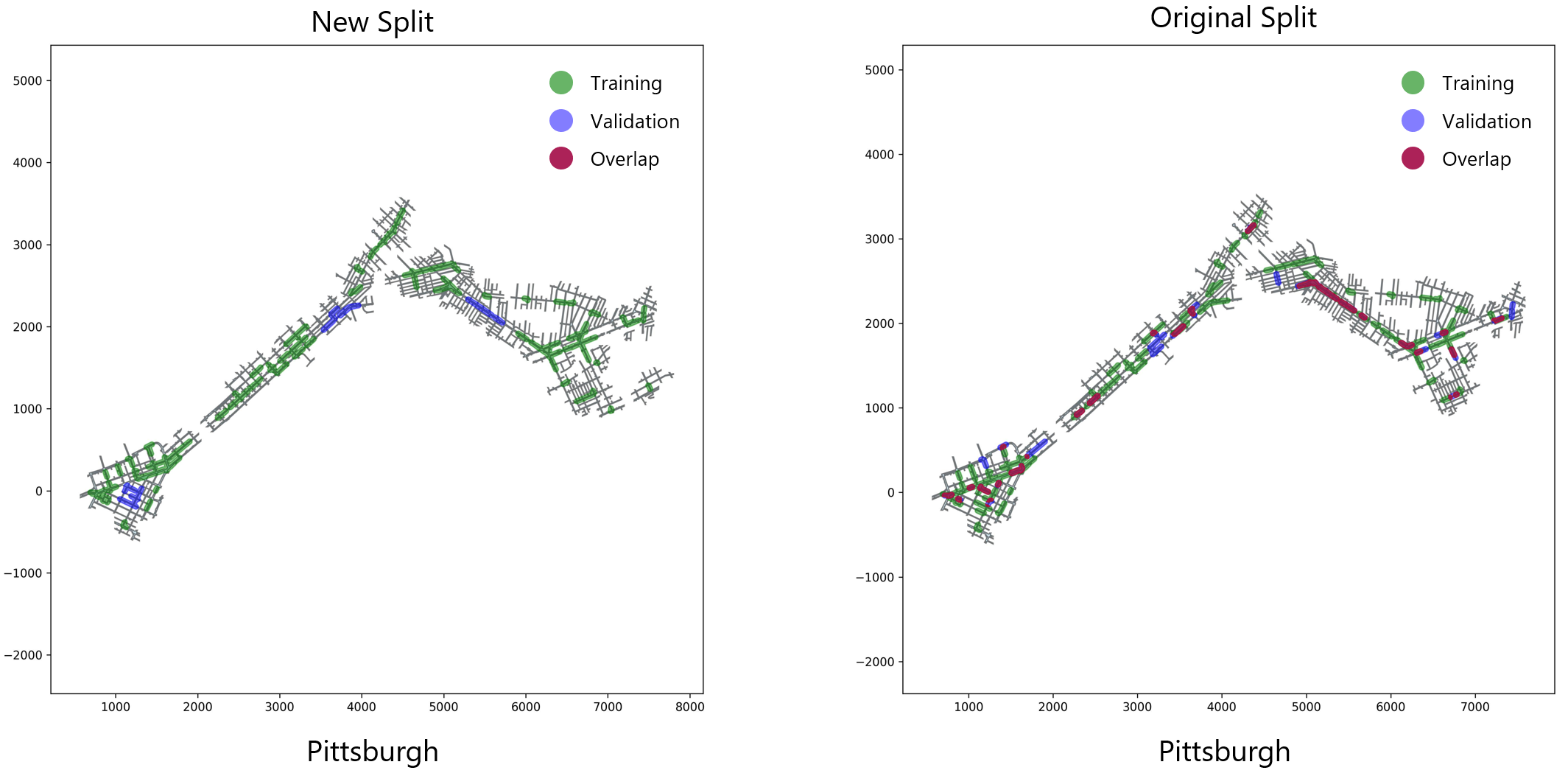}
   \caption{Comparison of the new and original splits in the Pittsburgh area of the Argoverse2 dataset. The \textit{green} regions represent the training set, \textit{blue} denotes the validation set, and \textit{red} signifies the overlapping areas. Best viewed in color.}
   \label{fig:PIT}
\end{figure*}

\begin{figure*}[t]
  \centering
   \includegraphics[width=1.0\linewidth]{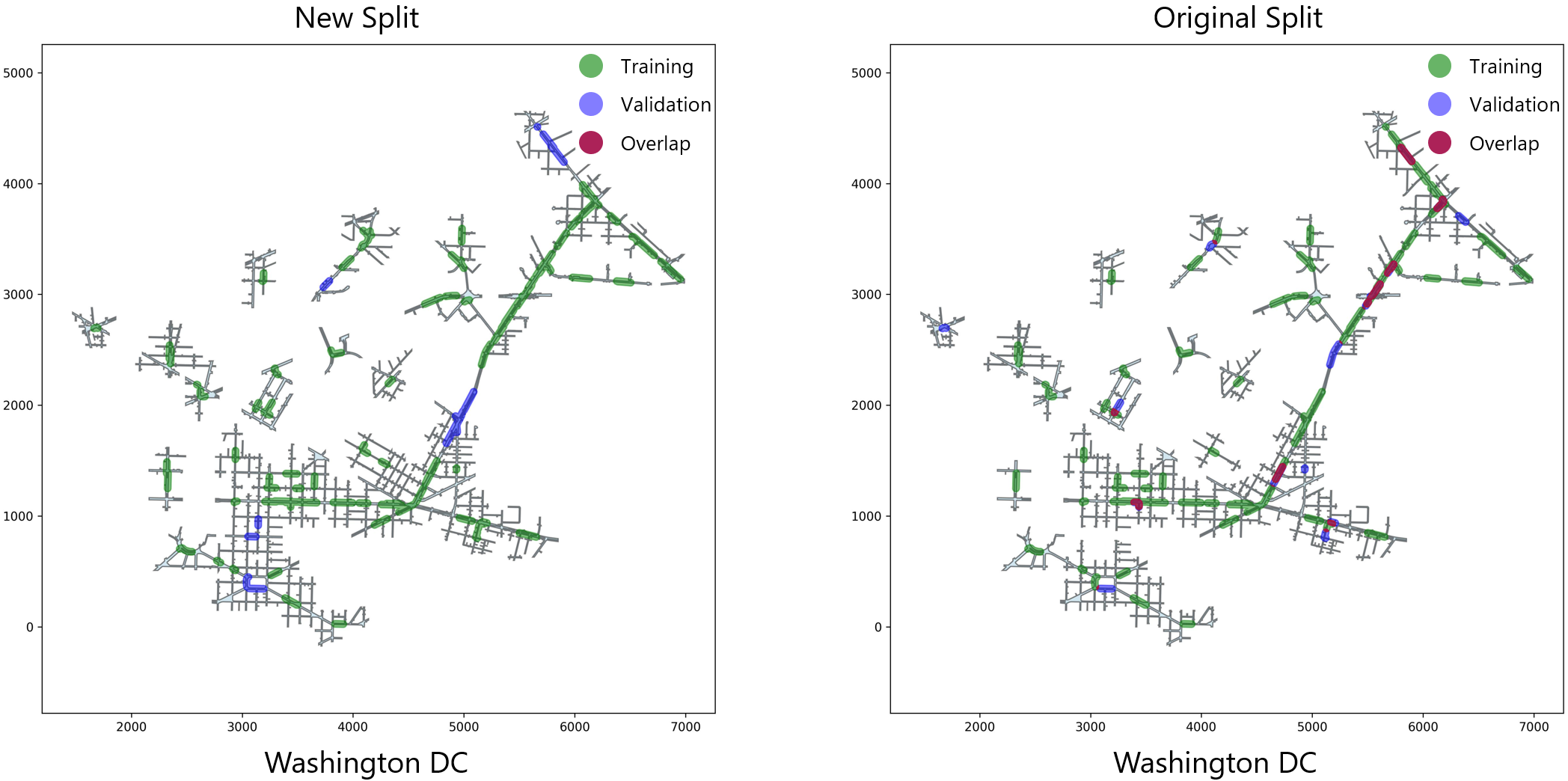}
   \caption{Comparison of the new and original splits in the Washington D.C. area of the Argoverse2 dataset. The \textit{green} regions represent the training set, \textit{blue} denotes the validation set, and \textit{red} signifies the overlapping areas. Best viewed in color.}
   \label{fig:WDC}
\end{figure*}

\begin{figure*}[t]
  \centering
   \includegraphics[width=1.0\linewidth]{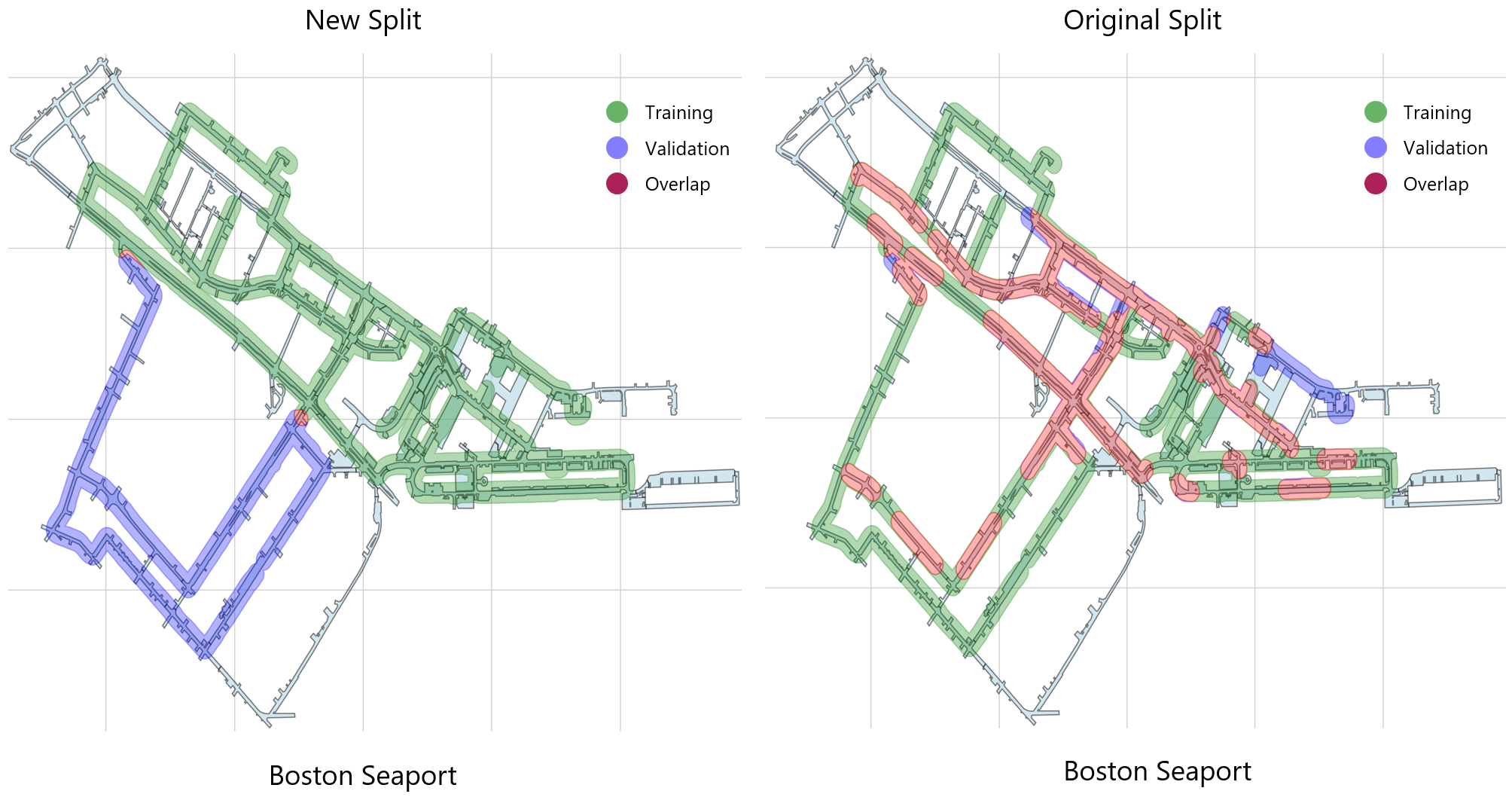}
   \caption{Comparison of the new and original splits in the Boston seaport area of the NuScenes dataset. The \textit{green} regions represent the training set, \textit{blue} denotes the validation set, and \textit{red} signifies the overlapping areas. Best viewed in color.}
   \label{fig:boston-seaport}
\end{figure*}

\begin{figure*}[t]
  \centering
   \includegraphics[width=1.0\linewidth]{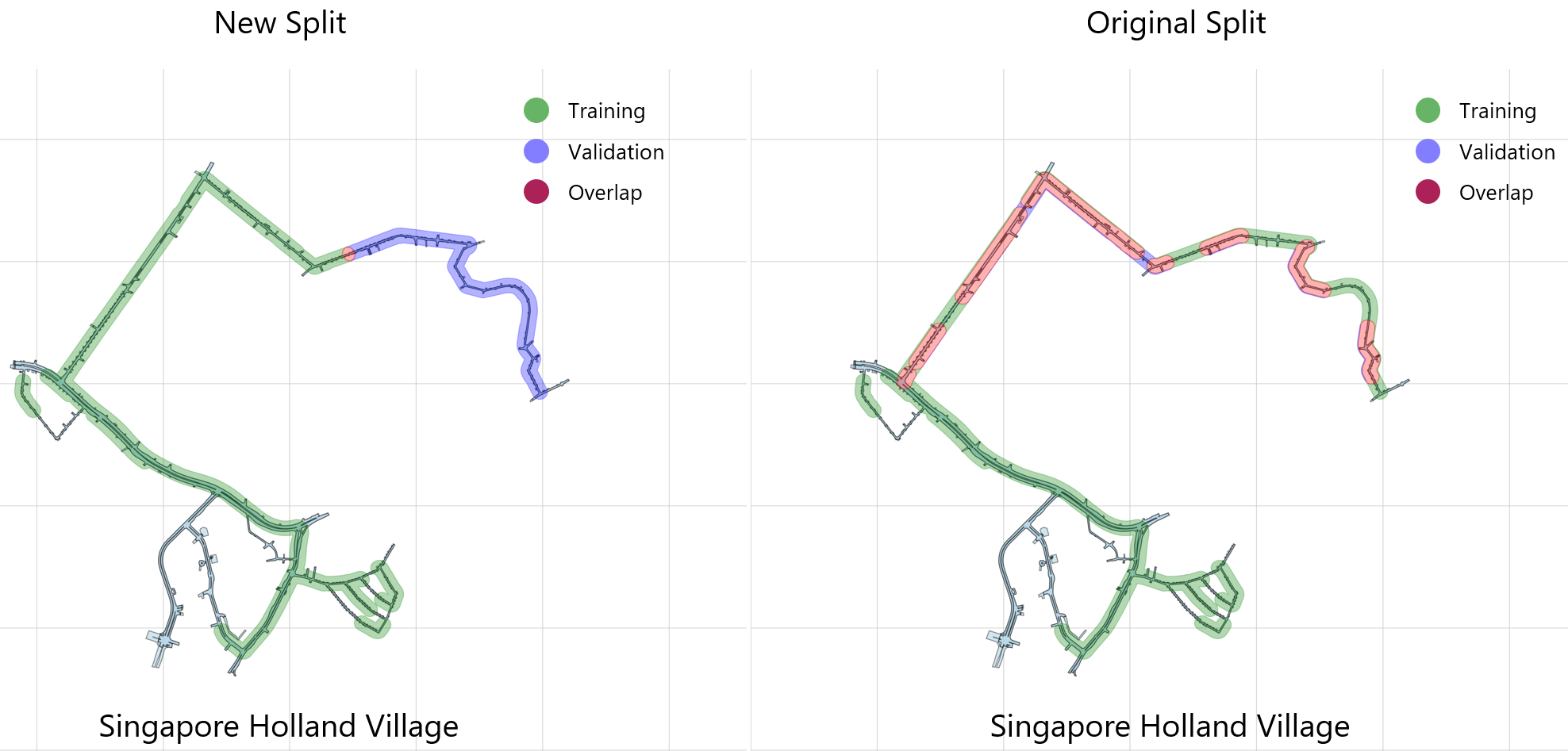}
   \caption{Comparison of the new and original splits in the Singapore Holland area of the NuScenes dataset. The \textit{green} regions represent the training set, \textit{blue} denotes the validation set, and \textit{red} signifies the overlapping areas. Best viewed in color.}
   \label{fig:singaport-holland}
\end{figure*}

\begin{figure*}[t]
  \centering
   \includegraphics[width=1.0\linewidth]{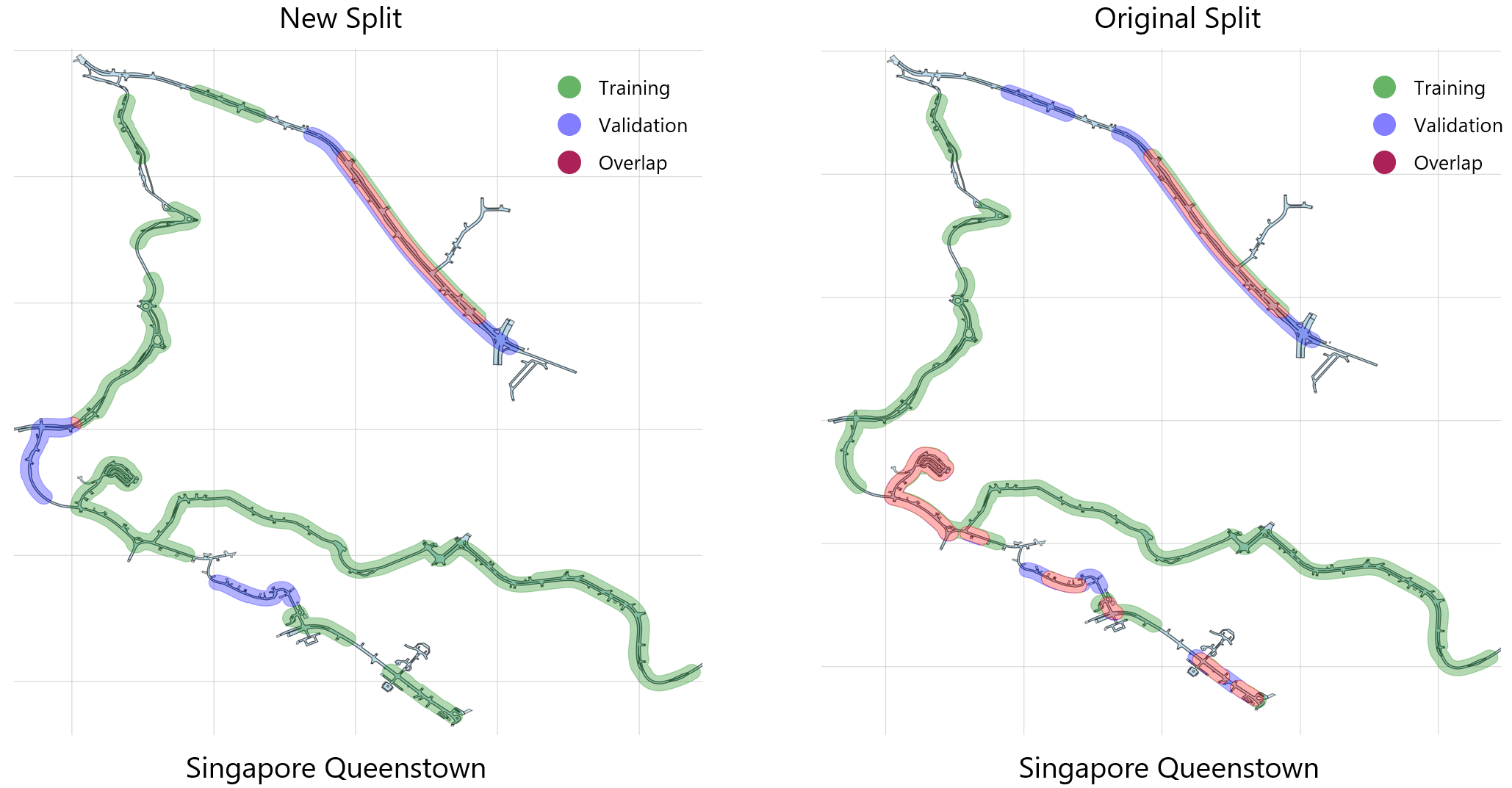}
   \caption{Comparison of the new and original splits in the Singapore Queenstown area of the NuScenes dataset. The \textit{green} regions represent the training set, \textit{blue} denotes the validation set, and \textit{red} signifies the overlapping areas. Best viewed in color.}
   \label{fig:singaport-queenstown}
\end{figure*}

\begin{figure*}[t]
  \centering
   \includegraphics[width=1.0\linewidth]{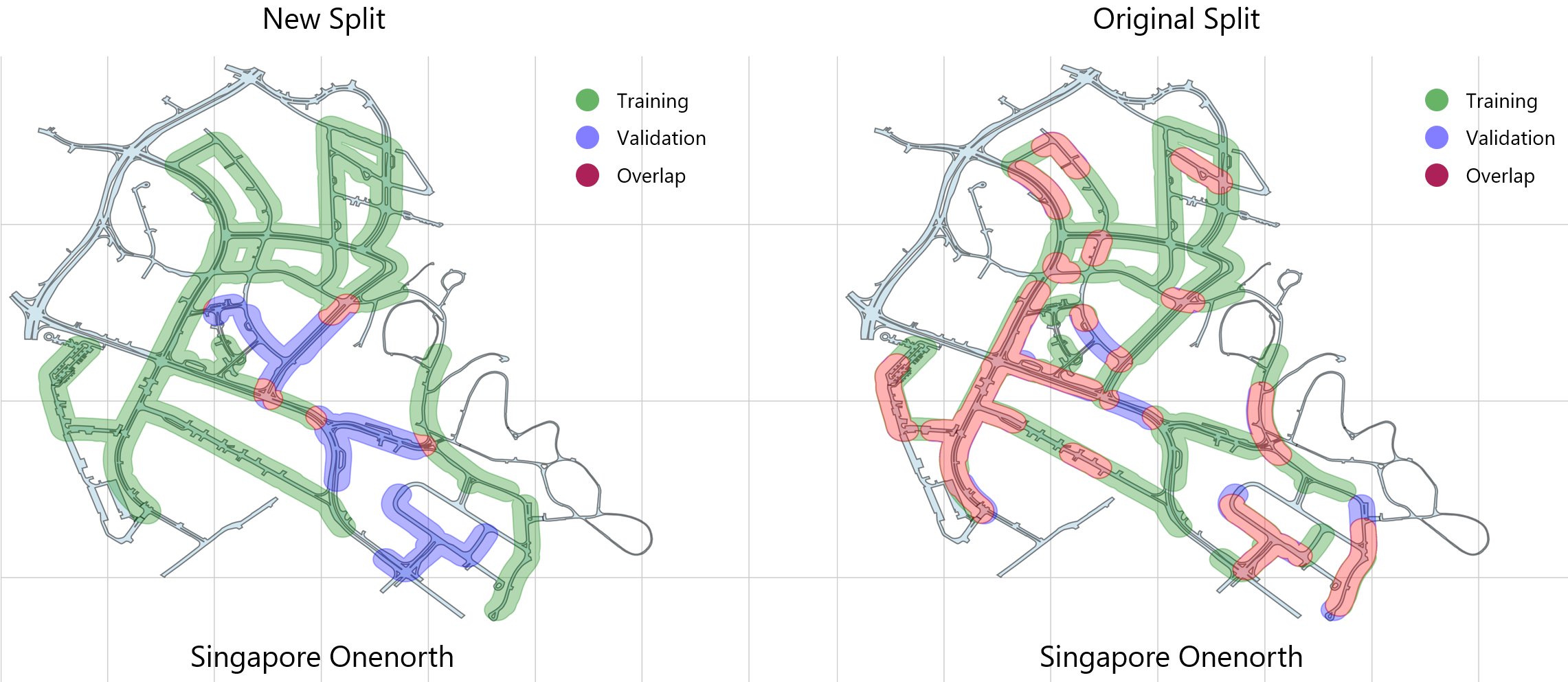}
   \caption{Comparison of the new and original splits in the Singapore One-North area of the NuScenes dataset. The \textit{green} regions represent the training set, \textit{blue} denotes the validation set, and \textit{red} signifies the overlapping areas. Best viewed in color.}
   \label{fig:singaport-onenorth}
\end{figure*}

To facilitate a more intuitive understanding, we visualize each city/district in both datasets, comparing the train/validation splits before and after the re-division. Figure~\ref{fig:ATX}, \ref{fig:DTW}, \ref{fig:MIA}, \ref{fig:PAO}, \ref{fig:PIT}, and \ref{fig:WDC} illustrate this comparison for Argoverse2. The \textit{green} areas represent the training set, \textit{blue} signifies the validation set, and \textit{red} indicates the overlaps. We maintain a balance between training and validation data across all six cities and ensure no overlaps. Figure~\ref{fig:boston-seaport}, \ref{fig:singaport-holland}, \ref{fig:singaport-queenstown}, and \ref{fig:singaport-onenorth} showcase the comparison for NuScenes. The original split results in a high degree of overlap, which the new split significantly alleviates.

Argoverse2 exhibits a significantly higher diversity of locations compared to NuScenes, which explains the performance gap observed in our experiments.

\end{document}